\documentclass[10pt,twocolumn,letterpaper]{article}

\usepackage[pagenumbers]{wacv}              % To produce the CAMERA-READY version
\usepackage[accsupp]{axessibility}
\definecolor{wacvblue}{rgb}{0.21,0.49,0.74}
\usepackage[pagebackref,breaklinks,colorlinks,allcolors=wacvblue]{hyperref}

\def\confName{WACV}
\def\confYear{2026}

\title{Multi-Receptive Field Ensemble with Cross-Entropy Masking for Class Imbalance in Remote Sensing Change Detection}

\author{
Humza Naveed$^{1,2}$ \quad
Xina Zeng$^{1,2}$ \quad
Mitch Bryson$^{1,2}$ \quad
Nagita Mehr Seresht$^{3}$\\[0.5em]
$^{1}$The University of Sydney  \hspace{2mm} $^{2}$ARIAM   \hspace{2mm}  $^{3}$Nearmap\\
{\tt\small
\{humza.naveed, xina.zeng, mitch.bryson\}@sydney.edu.au \quad
nagita.mehrseresht@nearmap.com
}
}

\begin{document}
\maketitle
\begin{abstract}
Remote sensing change detection (RSCD) is a complex task, where changes often appear at different scales and orientations. Convolutional neural networks (CNNs) are good at capturing local spatial patterns but cannot model global semantics due to limited receptive fields. Alternatively, transformers can model long-range dependencies but are data hungry, and RSCD datasets are not large enough to train these models effectively. To tackle this, this paper presents a new architecture for RSCD which adapts a segment anything (SAM) vision foundation model and processes features from the SAM encoder through a multi-receptive field ensemble to capture local and global change patterns. We propose an ensemble of spatial-temporal feature enhancement (STFE) to capture cross-temporal relations, a decoder to reconstruct change patterns, and a multi-scale decoder fusion with attention (MSDFA) to fuse multi-scale decoder information and highlight key change patterns. Each branch in an ensemble operates on a separate receptive field to capture finer-to-coarser level details. Additionally, we propose a novel cross-entropy masking (CEM) loss to handle class-imbalance in RSCD datasets. Our work outperforms state-of-the-art (SOTA) methods on four change detection datasets, Levir-CD, WHU-CD, CLCD, and S2Looking. We achieved 2.97\% F1-score improvement on a complex S2Looking dataset. The code is available at: \url{https://github.com/humza909/SAM-ECEM}  
\end{abstract}
    
\section{Introduction}\label{sc1}
Remote sensing change detection (RSCD) identifies changes between pairs of images captured by satellite or aerial imaging at different times. Correctly finding changes is crucial in applications such as future forecasting, environmental monitoring, land cover and land use analysis, and disaster management. RSCD methods have been studied extensively~\cite{changeformer,bit,changevit,ctdformer} with approaches ranging from simple feature differentiation~\cite{featurediff} to modern deep learning methods~\cite{changeformer, cgnet, fcsiam}. \\
\begin{figure}[t]
    \centering

       \includegraphics[width=0.5\textwidth]{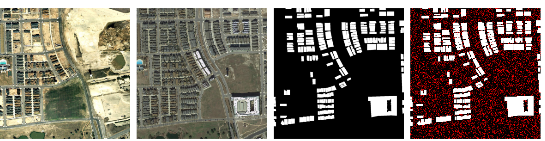} 
\caption{An illustration of dropped pixels in cross-entropy masking to handle class imbalance. The first three images are input images and ground truth, while the last image is an overlay with red pixels representing those dropped during loss calculation. }
    \label{fig:dropped}
\end{figure}
\begin{figure*}[t]
    \centering

       \includegraphics[width=1\textwidth]{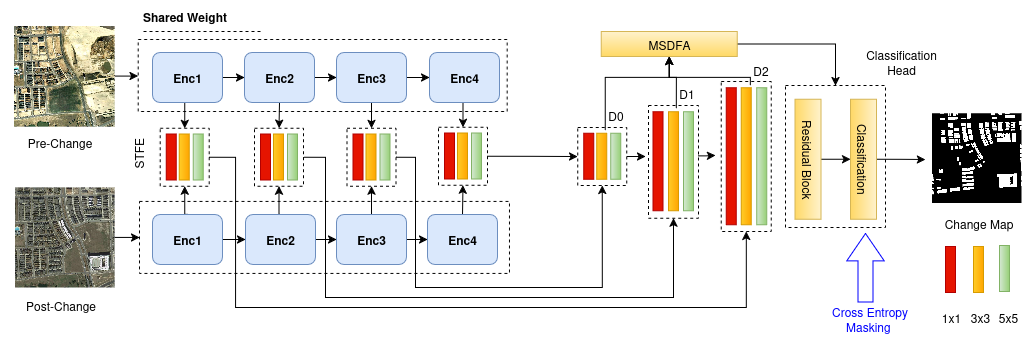} 
\caption{The architectural diagram of SAM-CEM-CD. The red, yellow, and green colored boxes represent an ensemble in spatio-temporal feature enhancement (STFE) and decoder, where each color operates on a separate receptive field from $1\times 1$, $3\times 3$, and $5\times 5$. The output of the red STFE boxes only connects with the red boxes in the decoder and so on.}
    \label{fig:architecture}
\end{figure*}
\noindent
Deep learning methods for change detection often employ pre-trained models as feature extractors. These pre-trained models are trained on large-scale vision datasets containing related tasks and domains. Because of this, the features extracted from these models are generalizable and transferable to other vision domains, such as change detection. ResNet~\cite{resnet} was used in bi-temporal image transformers for change detection (BIT) in~\cite{bit}, VGG~\cite{vgg} backbone was used in change-prior based change detection (CGNET)~\cite{cgnet}, and data-efficient transformers (deit)~\cite{deit} were used in Changevit~\cite{changevit}. Following this, the FastSAM encoder, a segment anything model (SAM) variant, was used in SAM-CD~\cite{samcd}. It demonstrated better performance over past approaches by using an encoder trained on a dense prediction task for change detection, which itself is a dense prediction task, compared to an encoder trained on non-dense prediction tasks. These backbones fine-tune well for remote sensing change detection, where dataset sizes are comparatively smaller compared to other domains, yielding significant performance gains with respect to training these models from scratch~\cite{bit,cgnet,changevit}.  \\
Changes in remote sensing data occur on different scales with diverse object sizes and orientations. It is crucial for change detection models to capture finer and coarser details to accurately model bi-temporal changes. Many of the past approaches process multi-scale information through multi-resolution feature maps in the decoder~\cite{samcd, changeformer, cgnet, eatder}, however, the decoder runs through a single receptive field size, limiting it to capture details at a single granularity level.   \\
Additionally, RSCD is a highly class-imbalanced problem, where pixels of change are significantly less frequent than non-change pixels. For example, Levir-CD~\cite{levircd} and WHU-CD~\cite{whucd} datasets have only 5\% change pixels in the entire dataset. This class imbalance introduces an inherent bias in the learning process in which models emphasize improving the accuracy of non-change areas over areas of change. Loss functions such as focal loss~\cite{focalloss}, dice loss~\cite{losses}, and weighted cross-entropy~\cite{losses} are suggested to handle class-imbalance. These loss functions adjust the importance of the majority and minority classes based on imbalance. However, the error for both classes propagates backwards, emphasizing more on the minority class that leads to overfitting and more false positives. \\
This paper presents a new RSCD architecture that addresses issues with class imbalance and local vs. global change information that are critical in change detection. We propose an architecture that uses an ensemble to learn features at different levels of granularity. Each branch in an ensemble operates on a different receptive field size to capture local and global details, which are combined for classification. We also propose a loss mechanism for class imbalance, which masks the loss calculation pixel-by-pixel for classes in the majority, as shown in Figure~\ref{fig:dropped}. Our architecture initially extracts multi-resolution features from a vision foundation model FastSAM~\cite{fastSAM} and learns cross-temporal relations using spatio-temporal feature enhancement (STFE), followed by a decoder ensemble to refine and upsample change patterns. The multi-scale decoder fusion with attention (MSDFA) fuses multi-scale information from the decoder and highlights important change regions for the classification head. \\
The key contributions of our work are:
\begin{itemize}
    \item We propose an ensemble of STFE, decoder, and MSDFA, where each branch operates on a range of receptive field sizes to capture local and global contextual changes at different scales. 
    \item We propose a novel cross-entropy masking (CEM) loss that randomly masks pixels for the majority class during loss calculation to reduce class imbalance for change detection, outperforming other class-imbalance losses. 
    \item We demonstrate the effectiveness of our architecture on four remote sensing change detection (RSCD) datasets, outperforming previous state-of-the-art (SOTA) methods in RSCD. 
\end{itemize}

\section{Related Work}
Architectures in RSCD are single encoder-decoder~\cite{fcsiam}, dual encoder single decoder~\cite{changeformer, cgnet}, or dual encoder-decoder~\cite{samcd} architectures, which are mainly divided into three categories: 1) early fusion, 2) middle fusion, and 3) late fusion. In early fusion~\cite{fcsiam}, temporal images are concatenated before processing through a deep learning model. In middle fusion~\cite{fcsiam,cgnet,eatder}, features from both encoder branches are combined to learn cross-temporal relationships, whereas in late fusion~\cite{samcd} features are processed through encoder-decoder branches and merged together at the end for change map classifcation. \\
Temporal images were concatenated in early convolutional neural networks (CNNs) based change detection methods~\cite{fcsiam}. The input data was processed through a UNet-style encoder-decoder architecture containing skip connections with multi-scale information processing. In another siamese-style variant for change detection by the same work~\cite{fcsiam}, features extracted at multi-resolutions are subtracted to highlight change features. However, simple concatenation of bi-temporal multi-resolution features in difference branch was shown to be better to learn change features~\cite{changeformer,snunet}. Instead of sparse connections, a densely connected UNet change detection architecture was proposed in~\cite{snunet}. Change priors were used to guide change detection in~\cite{cgnet}. This work also uses pre-trained backbones such as VGG~\cite{vgg} instead of training feature extractors from scratch. To emphasize key change regions, channel attention, spatial attention, or other attention mechanisms were used in~\cite{channel_attn_rscd,dasnet,spatial_attn_rscd}.   \\
The self-attention in transformers can capture global contextual dependencies opposite to CNNs. Based on this, many change detection methods were proposed using transformers~\cite{changevit,swinsunet,eatder}. A combination of CNNs and transformers were proposed in~\cite{bit, ctdformer}, extracting initial features from ResNet~\cite{resnet} followed by transformers-based encoder-decoder architecture.
ChangeFormer~\cite{changeformer} introduced a hierarchical transformer encoder to learn multi-resolution features similar to CNNs. However, transformers lack CNNs inductive bias and translational invariance, making them data hungry. Furthermore, the RSCD datasets~\cite{levircd,whucd} are not large enough to train models from scratch. \\
Because of this, researchers have started adopting pre-trained vision foundational models for RSCD. Generalized features extracted from these models are used to train robust change detection models. In~\cite{clipcd}, features from the remote-clip model are concatenated with the encoders in ChangeFormer~\cite{changeformer} and BIT~\cite{bit} models for better generalization. Segment anything model (SAM)~\cite{SAM} CNN-based variant FastSAM~\cite{fastSAM} was used in~\cite{samcd}. Features from the FastSAM encoder are used to reconstruct bi-temporal images through a decoder followed by late fusion and classification.    \\
Following this, our work employs the FastSAM vision foundation model for change detection. We use an ensemble of smaller and wider receptive fields to capture local and global structural changes in the data. Additionally, we propose a cross-entropy masking loss for a better generalization of the model. 
\section{Proposed Method}
\label{proposed_method}
\subsection{Siamese FastSAM Encoder}
FastSAM~\cite{fastSAM} is a smaller and faster instance segmentation model compared to segment anything~\cite{SAM}. It contains Yolov8-seg architecture trained on only 2\% of SAM's data, achieving equivalent performance.  
The FastSAM consists of encoder-decoder architecture, where the encoder encodes the input image into a compressed representation merging multi-scale features through a feature pyramid network. Due to dense image segmentation training, feature representations learned by the encoder are better transferable to change segmentation maps than the ResNet or vision transformer models trained for classification as in~\cite{cgnet,changevit}. Our method adapts the FastSAM encoder by fine-tuning it for change detection in a Siamese-style architecture with weight sharing. \\
Given a set of two remote sensing images $I_1$ and $I_2$ for change detection, we process the images through the encoder. We extract four feature maps $F_1$, $F_2$, $F_3$, and $F_4$ from the encoder at four scales $1/4$, $1/8$, $1/16$, and $1/32$. 

% The Yolov8-seg has a segmentation head additional to the detection head where the model's encoder has a CNN backbone followed by a feature pyramid network to integrate multi-scale features. The FastSAM backbone uses a sequence of cross-stage partial network (C2F) modules that split the input into two branches for better information and gradient flow. The last layer of the backbone consists of a spatial pyramid pooling fast (SPPF) module to fuse multi-scale features.

\subsection{Spatial-Temporal Feature Enhancement}
We concatenate features extracted at each scale to learn cross-temporal relations and differences in feature maps. Change patterns within a single resolution feature map occur at different sizes; therefore, it is important to learn spatio-temporal representations at different contextual levels. Contrary to the work in~\cite{changeformer, samcd,cgnet,snunet}, our spatial-temporal feature enhancement (STFE) module processes the input using an ensemble of convolutional receptive fields of sizes $1\times1$, $3\times3$, and $5\times5$ to model changes at various scales. We show the effectiveness of using a multi-receptive field ensemble compared to a single receptive field in Table~\ref{table_ablation_receptive_field}. The STFE module is replicated four times for different resolution feature maps, as shown in Figure~\ref{fig:architecture}.\\
Our spatial-temporal feature enhancement (STFE) module is represented by:
\begin{equation}
F_i^{STFE} = SiLU(BN(Conv2DWS_{k\times k}(F_j^{pre}, F_j^{post})))
\label{eq_diff}
\end{equation}
where $F_i^{STFE}$ are the enhanced feature maps, $F_j^{pre}$ and $F_j^{post}$ are pre-change and post-change images, respectively, $Conv2DWS$ is 2D depthwise separable convolution, $BN$ is batch normalization, $SiLU$ is sigmoid linear unit activation function, $i$ is equal to 1, 3, or 5 representing the convolutional receptive field, and $j$ represents the feature map output from the encoder.

\subsection{Decoder Ensemble}
%Multi-scale change features from the STFE module are used to reconstruct the change map in the UNet-style decoder~\cite{unet}. The feature map resolution in the decoder gradually increases with skip connections, concatenating features from the STFE with the decoder output. 
The decoder in our architecture is an ensemble of UNet-style decoders~\cite{unet}. Each decoder in ensemble operates in one of the convolutional receptive fields of $1\times1$, $3\times3$, and $5\times5$, and receives input from the $F_i^{STFE}$ output of the respective receptive field. This design of the model allows for emphasizing and reconstructing changes at different level of details, finer to coarse. The decoder block is written as:
\begin{equation}
\begin{aligned}
F_{up} &= ConvTrans2D_{2x2}(F_i^{STFE}) \\ 
F_{out} &= SiLU(BN(Conv2D_{i\times i}(F_{up}, F_{dec}))    
\label{eq_dec}
\end{aligned}
\end{equation}
In equation~\ref{eq_dec}, the feature map from the STFE branch $F_i^{STFE}$ is upsampled $F_{up}$ with transposed convolution $ConvTrans2D_{2x2}$ of kernel size $2\times2$ to match the spatial resolution of the decoder output. The $F_{up}$ is concatenated with the previous output of the decoder to process through the next decoder block, providing contextual information on changes in the bitemporal images. We empirically found that using standard 2D convolution in decoder performs better than depthwise separable convolution, which allows cross-channel interaction for better reconstruction. 
\subsection{Multi-Scale Decoder Fusion with Attention}
The early layers of the decoder contain rich global semantic information, whereas the last layers have local change structural detail. The decoder processes this information at multi-spatial resolutions, taking into account objects and changes in the temporal images at different scales. \\
Multi-scale decoder fusion with attention (MSDFA) concatenates the output of the decoder blocks, shown in equation~\ref{dec_concate}.
\begin{equation}
x_i = D_{2,i} \oplus upsample(D_{1,i}) \oplus upsample(D_{0,i})
\label{dec_concate}
\end{equation}
Upsample is a bilinear interpolation operation to match the feature resolution of the early decoder blocks to the last decoder block. The $i$ represents the decoder ensemble number based on the receptive field sizes 1, 3, or 5. Before attention, we inject local context using depthwise separable convolution on $x_i$ with a receptive field size same as the decoder. Afterwards, we apply channel attention followed by the spatial attention as given in equations~\ref{c_attn} and~\ref{a_attn}, respectively.
\begin{equation}
z_i = x_i \otimes \sigma ( Conv1D_{1 \times k} (GAP(x_i)) )
\label{c_attn}
\end{equation}
\begin{equation}
y_i = z_i \otimes \sigma ( Conv2D_{k \times k} (Avg(z_i)\oplus Max(z_i)) )
\label{a_attn}
\end{equation}
In equation~\ref{dec_concate},~\ref{c_attn} and~\ref{a_attn}, $\oplus$ represents concatenation, $\otimes$ is elementwise multiplication, $\sigma$ is sigmoid activation function, $GAP$ is global average pooling, $Avg$ and $Max$ are channel-wise average and maximums. To mention here, $k$ is equal to 3, 5, and 7 when $i$ is equal to 1, 3, and 5, respectively. We select a larger odd-numbered receptive field in attention modules compared to the original receptive field in the ensemble. This allows attending on a wider context to highlight key change areas and performs better than keeping a single kernel size for all $i$'s. \\
The output $y_i$ is summed with the input $x_i$, making it a residual connection and written as $y_i = y_i + x_i$. Our attention module is similar to the convolution block attention module (CBAM)~\cite{cbam}, except our channel attention uses only GAP and convolution instead of a multi-layer perceptron layer (MLP). Additionally, we have a separate MSDFA for each decoder in the decoder ensemble. The final output from the MSDFA is given as:
\begin{equation}
y = y_1 \oplus y_3 \oplus y_5
\end{equation}

\subsection{Classification Head}
The last layer in our architecture is the classification head to generate change maps. It comprises six residual layers followed by a convolutional classification layer. The input to the classification head is the concatenated output from the MSDFA layer.  
\subsection{Cross-Entropy Masking for Class Imbalance}
The bitemporal change detection datasets are heavily imbalanced in change and unchanged regions. For instance, Levir-CD~\cite{levircd} and WHU-CD~\cite{whucd} have 95\% unchanged pixels and only 5\% change pixels. Because of this imbalance, the cross-entropy loss in change detection is highly biased towards accurately classifying unchanged pixels instead of change pixels. To ensure that the model emphasizes change pixels as equally as unchanged pixels, we randomly dropout the unchanged pixels from the binary cross-entropy loss. For this, we compute the per-element binary cross-entropy (BCE) as mentioned:

\begin{equation}
L_{bce} = -y_i \log(\hat{y_i}) + (1 - y_i) \log(1 - \hat{y_i})
\label{eq_bce}
\end{equation}
where $y_i$ is the ground-truth and $\hat{y_i}$ is the predicted label. To dropout the pixels in binary cross-entropy, we create a mask by keeping all the loss values for change pixels and dropping values for the unchanged pixels based on the dropout value. This can be written as follows: 
\begin{equation}
M_i = 
\begin{cases} 
1 & \text{if } y_i = 1 \\ 
1 & \text{if } y_i = 0 \text{ and } (R_i > \delta)\\
0 & \text{if } y_i = 0 \text{ and } (R_i < \delta)
\end{cases}
\label{eq_mask}
\end{equation}

In equation~\ref{eq_mask}, $R_i$ is a matrix of the same dimensions as the change map with values sampled from the Uniform distribution function i.e. $R_i \sim \text{Uniform}(0,1)$. The $\delta$ in equation~\ref{eq_mask} is the dropout probability that defines the number of unchanged pixels to drop from loss calculation. The overall binary cross-entropy loss function with dropout becomes:
\begin{equation}
L_{drop} = \frac{\sum_{i=1}^{N} L_{bce} \cdot M_i}{\sum_{i=1}^{N} M_i}
\end{equation}

\begin{table*}[t]
    \centering
    \caption{Quantitative results of different RSCD methods on the Levir-CD and WHU-CD datasets, results are in \%. The best results are highlighted.}
    \resizebox{0.8\linewidth}{!}{%
        \begin{tabular}{r|ccccc|ccccc}
        \toprule
            Method & \multicolumn{5}{c}{Levir-CD} & \multicolumn{5}{c}{WHU-CD} \\
            \cline{2-11}
            & $Pre$ & $Rec$ & $OA$ & $mF_{1}$ & $mIoU$ & $Pre$ & $Rec$ & $OA$ & $mF_{1}$ & $mIoU$ \\
            \hline
            FC-Siam-diff\cite{fcsiam} & 92.93 & 87.02 & 98.15 & 89.73 & 82.67 & 94.52 & 89.32 & 98.70 & 91.74 & 85.65\\
            FC-Siam-conc\cite{fcsiam} & 92.03 & 89.82 & 98.28 & 90.89 & 84.34 & 93.94 & 92.83 & 98.90 & 93.37 & 88.19\\
            SNUNet\cite{snunet} & 93.83 & 90.11 & 98.50 & 91.88 & 85.84 & 92.61 & 83.26 & 98.10 & 87.30 & 79.40\\
            BIT\cite{bit} & 90.27 & 83.37 & 97.60 & 86.46 & 78.23 & 87.92 & 93.41 & 98.29 & 90.46 & 83.72\\
            ChangeFormer\cite{changeformer} & 91.46 & 86.31 & 97.95 & 88.69 & 81.21 & 96.28 & 92.95 & 99.12 & 94.55 & 90.08\\
            CTD-Former\cite{ctdformer} & 93.60 & 91.85 & 98.62 & 92.71 & 87.11 & 96.74 & 97.03 & 99.50 & 96.89 & 94.11\\
            CGNet\cite{cgnet} & 95.95 & 94.95 & 99.13 & 95.44 & 91.58 & \textbf{98.04} & 96.14 & 99.52 & 97.07 & 94.44\\
            %Changer & 96.31 & 95.08 & 99.18 & 95.69 & 92.00 & \\
            EATDer\cite{eatder} & \textbf{96.29} & 92.70 & 98.87 & 94.41 & 89.84 & 95.57 & 93.06 & 99.01 & 94.28 & 89.64 \\
            SAM-CD~\cite{samcd} & 95.87 & 95.14 & 99.14 & 95.50 & 91.68 & 96.91 & 95.42 & 99.36 & 96.15 & 92.81 \\
            \hline
            SAM-ECEM (ours) & 96.27 & \textbf{95.72} & \textbf{99.23} & \textbf{96.01} & \textbf{92.56} & 97.82 & \textbf{96.69} & \textbf{99.54} & \textbf{97.25} & \textbf{94.76} \\
            
            %& \textbf{99.50} & \textbf{97.02} & \textbf{94.35}
        \bottomrule
        \end{tabular}
        }\label{table_levir_whu}
\end{table*}
\section{Experimentation and Results}
\label{results}

\begin{table}[!htbp]
    \centering
    \caption{Quantitative results of different RSCD methods on the CLCD dataset, results are in \%.}
    \resizebox{0.9\linewidth}{!}{%
        \begin{tabular}{r|ccccc}
        \toprule
            Method & \multicolumn{5}{c}{Accuracy Metrics}  \\
            \cline{2-6}
            & $Pre$ & $Rec$ & $OA$ & $mF_{1}$ & $mIoU$ \\
            \hline
            FC-Siam-diff\cite{fcsiam} & 83.13 & 74.60 & 94.73 & 78.10 & 68.19 \\
            FC-Siam-conc\cite{fcsiam} & 81.65 & 79.94 & 94.84 & 80.77 & 70.98  \\
            SNUNet\cite{snunet} & 85.92 & 82.62 & 95.84 & 84.19 & 75.11  \\
            BIT\cite{bit} & 81.42 & 70.27 & 94.17 & 74.39 & 64.51  \\
            ChangeFormer\cite{changeformer} & 84.41 & 81.36 & 95.47 & 82.80 & 73.40 \\
            CTD-Former\cite{ctdformer} & 87.29 & 83.17 & 96.12 & 85.08 & 76.24\\
            CGNet\cite{cgnet} & 85.87 & 85.30 & 96.00 & 85.58 & 76.83 \\
            EATDer\cite{eatder} & 87.16 & 77.18 & 93.84 & 81.16 & 71.18 \\
            %Changer & & & & & & 73.15 & 61.05 & 99.26 & 66.55 & 74.56\\
            SAM-CD~\cite{samcd} & 88.25 & 85.65 & 96.26 & 86.89 & 78.53\\
            \hline
            SAM-ECEM (ours) & \textbf{88.43} & \textbf{87.53} & \textbf{96.73} & \textbf{88.04} & \textbf{80.14}\\
            
            %& \textbf{99.50} & \textbf{97.02} & \textbf{94.35}
        \bottomrule
        \end{tabular}
        }\label{table_CLCD}
\end{table}

\begin{table}[t]
    \centering
    \caption{Quantitative results of different RSCD methods on the S2Looking dataset for change class, results are in \%.}
    \resizebox{0.9\linewidth}{!}{%
        \begin{tabular}{r|ccccc|ccccc}
        \toprule
            Method & \multicolumn{4}{c}{Accuracy Metrics} \\
            \cline{2-5}
            & $Pre$ & $Rec$ & $F_{1}$ & $IoU$ \\
            \hline
            FC-Siam-diff\cite{fcsiam} & 83.49 & 32.32 & 46.60 & 30.38 \\
            FC-Siam-conc\cite{fcsiam} & 68.27 & 18.52 & 13.54 & - \\
            SNUNet\cite{snunet} & 45.26 & 50.60 & 47.78 & 31.39 \\
            BIT\cite{bit} &  70.26 & 56.53 & 62.65 & 45.62 \\
            ChangeFormer\cite{changeformer} & 72.82 & 56.13 & 63.39 & - \\
            CGNet\cite{cgnet} & 70.18 & 59.38 & 64.33 & 47.41\\
            EATDer\cite{eatder} & 65.85 & 54.74 & 59.78 & 42.64\\
            %Changer & & & & & & 73.15 & 61.05 & 99.26 & 66.55 & 74.56\\
            SAM-CD~\cite{samcd} & 72.80 & 58.92 & 65.13 & 48.29\\
            \hline
            SAM-ECEM (ours) & \textbf{74.59} & \textbf{62.65} & \textbf{68.10} & \textbf{51.63}\\
            %& \textbf{99.50} & \textbf{97.02} & \textbf{94.35}
        \bottomrule
        \end{tabular}
        }\label{table_S2Look}
\end{table}

\subsection{Implementation Details}
We used PyTorch for model training. We trained the model for 35 epochs with three additional warm-up epochs using a stochastic gradient descent (SGD) optimizer. We use the FastSAM encoder and fine-tune its pre-trained weights. The learning rate (lr) is 0.01 for Levir-CD and CLCD, and 0.001 for WHU-CD and S2Looking. We decay the learning rate in each iteration $lr*(1- iterations/50)^{decay}$, where decay is 2.0 for Levir-CD and CLCD, and 3.0 for WHU-CD and S2Looking. Image flipping and random cropping augmentations are applied during training and images are flipped 4 times for inference to produce stable results. More details can be seen in our GitHub repo: \url{https://github.com/humza909/SAM-ECEM}.
\subsection{Change Detection Datasets}
The following datasets are used for training and testing of our change detection framework:

\vspace{2mm}\noindent\textbf{Levir-CD~\cite{levircd}:}
This dataset contains 637 image pairs with 31,333 annotated change objects. The data is collected from Google Earth for six Texas cities between 2012 and 2016. It has 445 training, 64 validation, and 128 testing images.

\vspace{2mm}\noindent\textbf{WHU-CD~\cite{whucd}:}
This is an aerial imaging dataset captured over Christchurch, New Zealand, after the 2011 earthquake, between 2012-2016. We used the dataset split of 6096 training, 762 validation, and testing images provided by~\cite{samcd}.

\vspace{2mm}\noindent\textbf{CLCD~\cite{clcd}:}
The cropland change detection dataset is captured in Guangdong, China between 2017 to 2019. It consists of 600 image pairs collected by the Gaofen-2 satellite with a split of 320 training, 120 validation and testing image pairs.

\vspace{2mm}\noindent\textbf{S2Looking~\cite{s2looking}:}
This is a large-scale change detection dataset with 5000 image pairs collected from satellites such as Gaofen, SuperView, and BeiJing-2. The dataset presents various challenges, including misalignment and varying poses. 
\subsection{Evaluation Metrics}
To evaluate the performance of our model, we employ widely used metrics for change detection, including precision (Prec), recall (Rec), $F_1$ score, overall accuracy (OA), and intersection over union (IoU). Tables~\ref{table_levir_whu} and~\ref{table_CLCD} contain the average of F1, represented as mF1, and other metrics for the background and foreground classes, whereas in Table~\ref{table_S2Look}, the results are reported only for the foreground class. 
%\begin{align}
%P   &= \frac{TP}{TP + FP} \\
%R   &= \frac{TP}{TP + FN} \\
%F_1 &= 2 \cdot \frac{P \cdot R}{P + R} \\ 
%OA  &= \frac{TP + TN}{TP + FP + TN + FN} \\
%IoU &= \frac{TP}{TP + FP + FN}
%\end{align}
\begin{figure*}[t]
    \centering
    \setlength{\tabcolsep}{1pt} % Reduce horizontal spacing between images
    \renewcommand{\arraystretch}{0.5} % Reduce vertical spacing between rows
\resizebox{1.0\textwidth}{!}{
    \begin{tabular}{cccccccc}
        % Add header row with labels
       \fontsize{24}{16}\selectfont\textbf{T1} &
\fontsize{24}{16}\selectfont\textbf{T2} &
\fontsize{24}{16}\selectfont\textbf{GT} &
\fontsize{24}{16}\selectfont\textbf{CF} &
\fontsize{24}{16}\selectfont\textbf{BIT} &
\fontsize{24}{16}\selectfont\textbf{CGNet} &
\fontsize{24}{16}\selectfont\textbf{SAM-CD} &
\fontsize{24}{16}\selectfont\textbf{Ours} \\
        % Optionally add some vertical space after header
        %\\[-5pt] 
        
        % Your original image rows

        \includegraphics[width=0.3\textwidth]{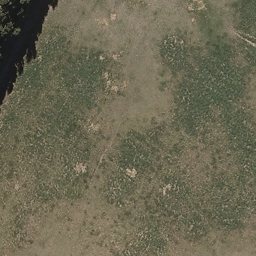} &
        \includegraphics[width=0.3\textwidth]{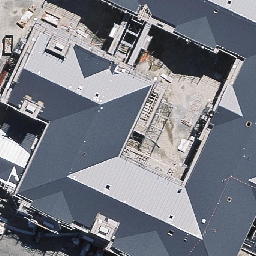} &
        \includegraphics[width=0.31\textwidth]{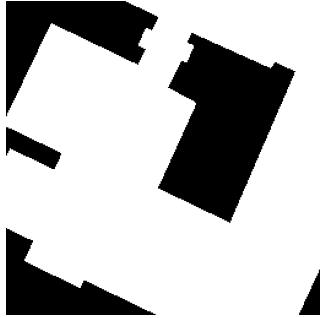} &
        \includegraphics[width=0.3\textwidth]{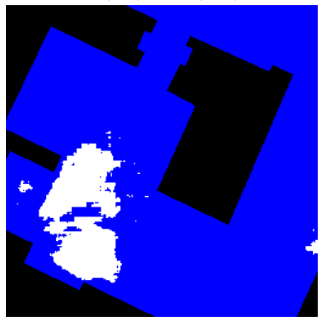} &
        \includegraphics[width=0.3\textwidth]{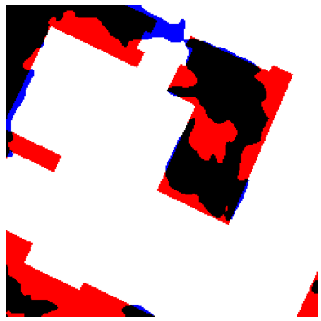} &
        \includegraphics[width=0.3\textwidth]{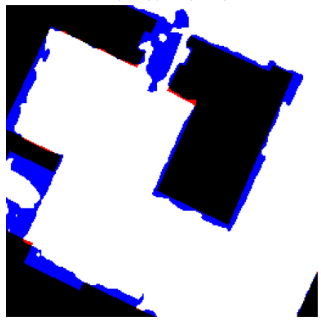} &
        \includegraphics[width=0.3\textwidth]{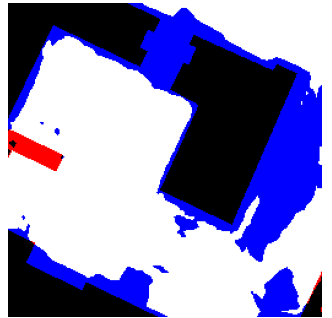} &
        \includegraphics[width=0.3\textwidth]{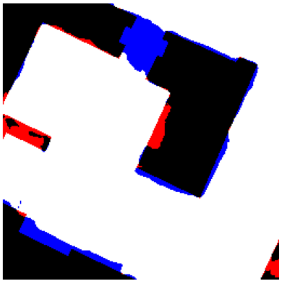} \\
     
        \includegraphics[width=0.3\textwidth]{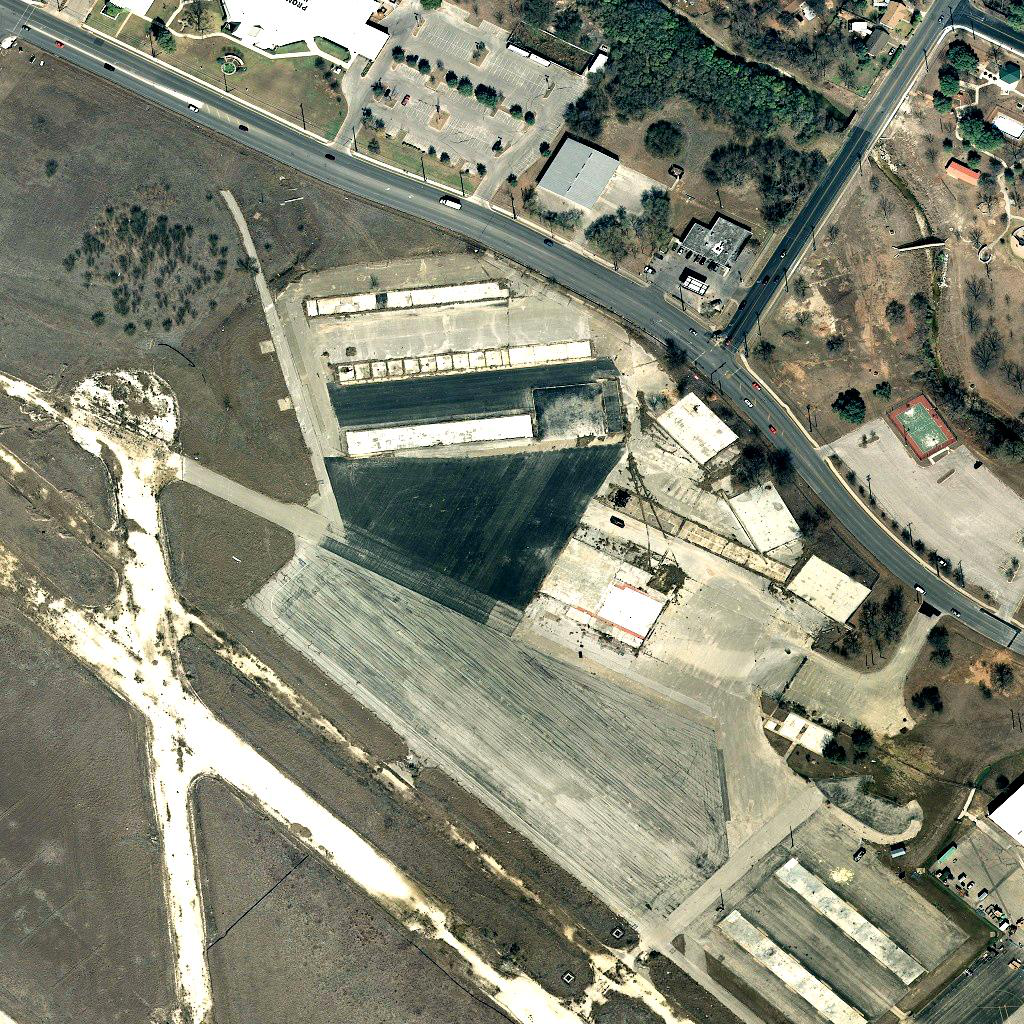} &
        \includegraphics[width=0.3\textwidth]{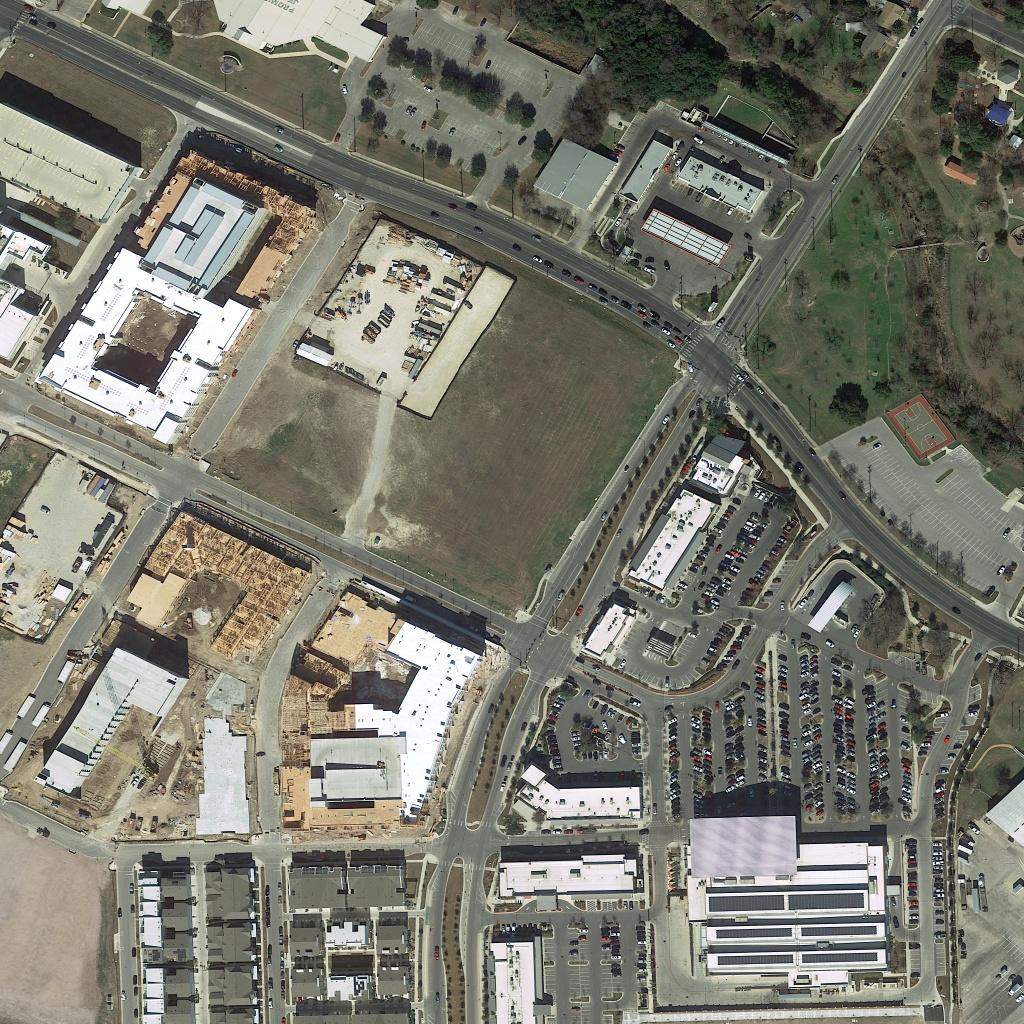} &
        \includegraphics[width=0.307\textwidth]{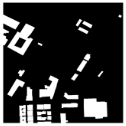} &
         \includegraphics[width=0.3\textwidth]{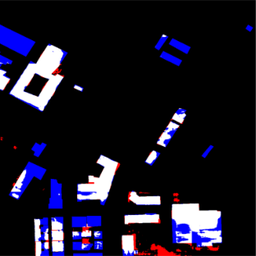} &
        \includegraphics[width=0.3\textwidth]{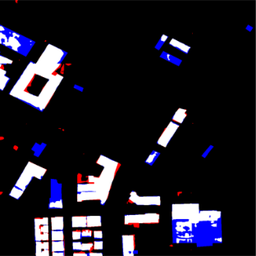} &
        \includegraphics[width=0.3\textwidth]{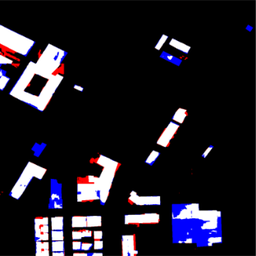} &
        \includegraphics[width=0.305\textwidth]{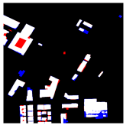} &
        \includegraphics[width=0.3\textwidth]{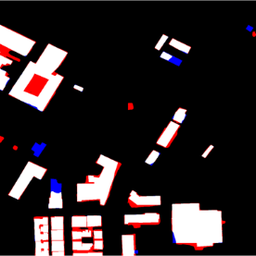} \\

       \includegraphics[width=0.3\textwidth,height=0.2\textheight]{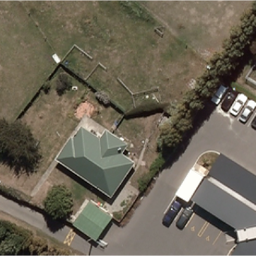} &
\includegraphics[width=0.3\textwidth,height=0.2\textheight]{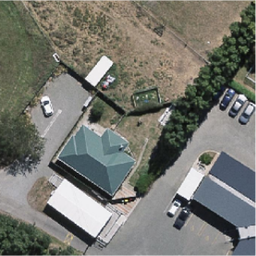} &
\includegraphics[width=0.3\textwidth,height=0.2\textheight]{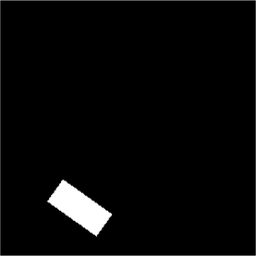} &
\includegraphics[width=0.3\textwidth,height=0.2\textheight]{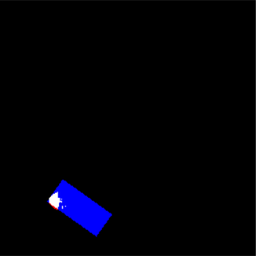} &
\includegraphics[width=0.3\textwidth,height=0.2\textheight]{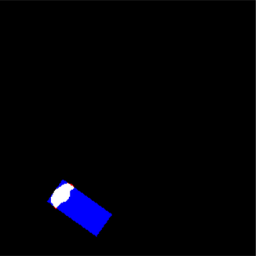} &
\includegraphics[width=0.3\textwidth,height=0.2\textheight]{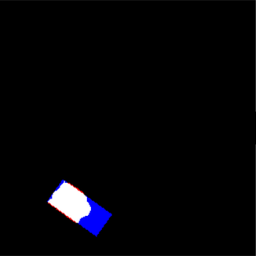} &
\includegraphics[width=0.3\textwidth,height=0.2\textheight]{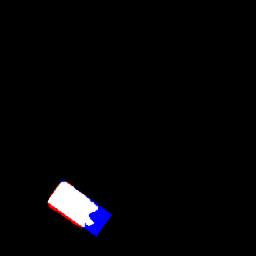} &
\includegraphics[width=0.3\textwidth,height=0.2\textheight]{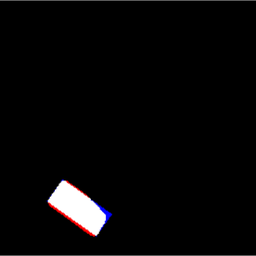} \\

\includegraphics[width=0.3\textwidth,height=0.2\textheight]{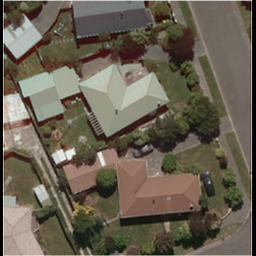} &
\includegraphics[width=0.30\textwidth,height=0.2\textheight]{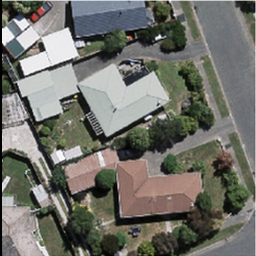} &
\includegraphics[width=0.3\textwidth,height=0.2\textheight]{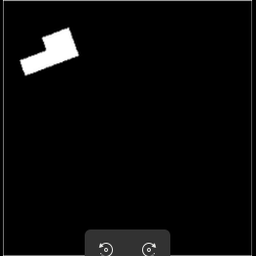} &
\includegraphics[width=0.3\textwidth,height=0.2\textheight]{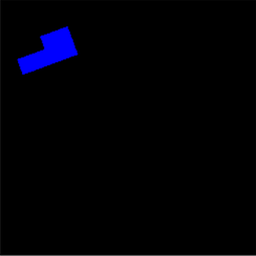} &
\includegraphics[width=0.3\textwidth,height=0.2\textheight]{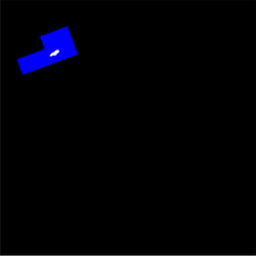} &
\includegraphics[width=0.3\textwidth,height=0.2\textheight]{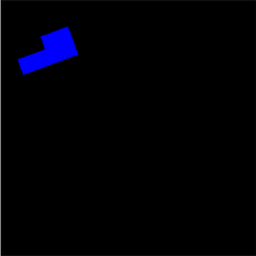} &
\includegraphics[width=0.3\textwidth,height=0.2\textheight]{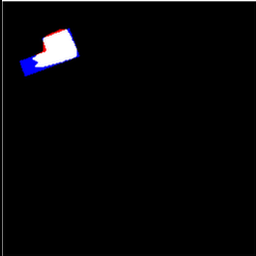} &
\includegraphics[width=0.3\textwidth,height=0.2\textheight]{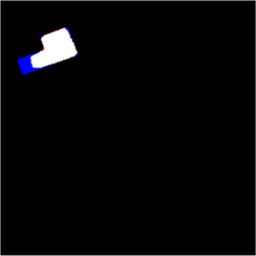} \\

\includegraphics[width=0.3\textwidth,height=0.2\textheight]{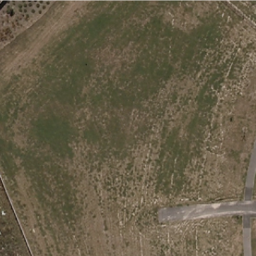} &
\includegraphics[width=0.3\textwidth,height=0.2\textheight]{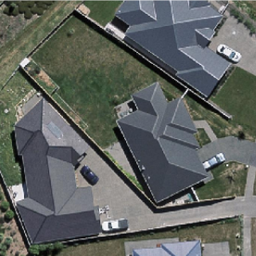} &
\includegraphics[width=0.3\textwidth,height=0.2\textheight]{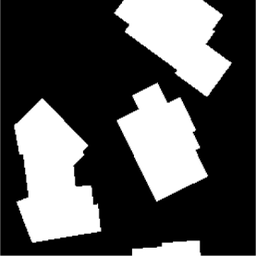} &
\includegraphics[width=0.3\textwidth,height=0.2\textheight]{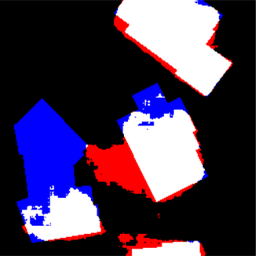} &
\includegraphics[width=0.3\textwidth,height=0.2\textheight]{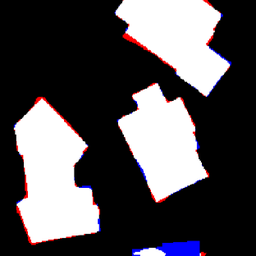} &
\includegraphics[width=0.3\textwidth,height=0.2\textheight]{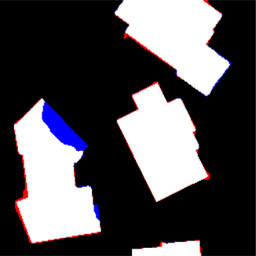} &
\includegraphics[width=0.3\textwidth,height=0.2\textheight]{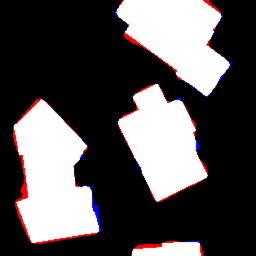} &
\includegraphics[width=0.3\textwidth,height=0.2\textheight]{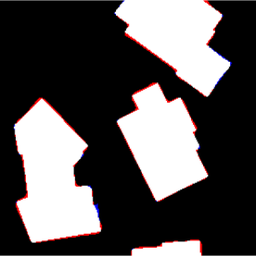} \\

    \end{tabular}}
    
    \caption{Comparison of change detection results. T1 and T2 are the input images at different timestamps, GT is the ground truth, and CF~\cite{changeformer}, BIT~\cite{bit}, CGNet~\cite{cgnet}, SAM-CD~\cite{samcd}, and SAM-ECEM (ours) are predictions from models. Red pixels are FPs and blue pixels are FNs.}
    \label{fig:comparison}
\end{figure*}
\subsection{Quantitative Results}
Our method, SAM-ECEM, achieves superior performance than other SOTA methods in the literature. For the correctness of the results, we follow the same evaluation strategy as in~\cite{samcd}. The results in Tables~\ref{table_levir_whu},~\ref{table_CLCD}, and~\ref{table_S2Look} are also collected from~\cite{samcd}. Table~\ref{table_levir_whu} shows an increase of 0.51\% and 0.88\% in mF1 and mIoU, respectively, for the Levir-CD dataset. Similarly, we achieve 97.25\% mF1 and 94.76\% mIoU for the WHU-CD dataset, compared to the second-best CGNET~\cite{cgnet}, which yields 97.07\% mF1 and 94.44\% mIoU. We recalculated the SAM-CD results for WHU-CD as given in Table~\ref{table_levir_whu}. Table~\ref{table_CLCD} shows an approximately 1.5\% improvement in mF1 and mIoU for CLCD against other methods, whereas for S2Looking, we achieve 2.97\% higher F1  and 3.34\% higher IoU performance than the SOTA methods in Table~\ref{table_S2Look}.

\subsection{Visualization}
We visually compare the change maps of ChangeFormer~\cite{changeformer}, BIT~\cite{bit}, CGNeT~\cite{cgnet}, SAM-CD~\cite{samcd} and SAM-ECEM (ours) in Figure~\ref{fig:comparison}. The blue predictions are FNs and the red predictions are FPs. We can see the reduced FNs and FPs by SAM-ECEM in the figure, which shows the significance of our method over other SOTA methods. 
\subsection{Ablation Study}
In this section, we examine the impact of various architectural components on performance.

\begin{table}[t]
    \centering
    \caption{Ablation study using different modules in our architecture on the Levir-CD dataset, results are in \%. STFE stands for spatio-temporal feature enhancement, D stands for Unet-decoder, MSDFA stands for multi-scale decoder fusion with attention, and CEM stands for cross-entropy masking.}
    \resizebox{0.8\linewidth}{!}{%
        \begin{tabular}{r|cccc}
        \toprule
            Method & $mF_{1}$ & $mIoU$ \\
        \midrule
            SAM + STFE + D & 95.83 & 92.24 \\
            SAM + STFE + D + MSDFA & 95.88 & 92.33 \\
            SAM + STFE + D + MSDFA + CEM & \textbf{96.01} & \textbf{92.56} \\
            
        \bottomrule
        \end{tabular}
    }
    \label{table_ablation}
\end{table}
\vspace{2mm}\noindent\textbf{Architectural Modules:}
We compared the impact of various modules in our architecture. We demonstrate the improvement in performance over the baseline by adding MSDFA in Table~\ref{table_ablation}, which fuses multi-scale information from the reconstructed change maps. The mF1 increases from 95.83\% to 95.88\%, and mIoU increases from 92.24\% to 92.33\% by using MSDFA. Training the network with CEM loss improves performance and generalizability, increasing mF1 from 95.88\% to 96.01\%, and mIoU from 92.33\% to 92.56\%.

\begin{table}[t]
    \centering
    \caption{Ablation study on using different receptive field sizes in the model, results are for Levir-CD dataset.}
    \resizebox{0.8\linewidth}{!}{%
        \begin{tabular}{r|cccc}
        \toprule
            Method & $mF_{1}$ & $mIoU$ \\
        \midrule
            Receptive Field = 1 & 95.94 & 92.43 \\
            Receptive Field = 3 & 95.92 & 92.39 \\
            Receptive Field = 5 & 95.96 & 92.46 \\
            Ensemble of Receptive Field = 1, 3, 5 & \textbf{96.01} & \textbf{92.56} \\
            
        \bottomrule
        \end{tabular}
    }
    \label{table_ablation_receptive_field}
\end{table}
\vspace{2mm}\noindent\textbf{Ensemble of Receptive Fields:}
We analyze the effect of using an ensemble of different receptive field sizes. The analysis in Table~\ref{table_ablation_receptive_field} shows that using a combination of receptive fields $1\times 1$, $3 \times 3$, and $5 \times 5$ is better at capturing local and global semantics than using a single branch of one of these receptive fields. All of these experiments are performed using CEM loss. 

\begin{table}[t]
    \centering
    \caption{Ablation study on using different kernel sizes in MSDFA for different branches in an ensemble, results are for Levir-CD dataset. $k$ is the kernel size in MSDFA, and $i$ is the receptive field size in the ensemble. }
    \resizebox{0.7\linewidth}{!}{%
        \begin{tabular}{r|cccc}
        \toprule
            Method & $mF_{1}$ & $mIoU$ \\
        \midrule
            MSDFA (k=3, i=1,3,5) & 95.99 & 92.52 \\
            MSDFA (k=5, i=1,3,5) & 95.99 & 92.53 \\
            MSDFA (k=7, i=1,3,5) & 95.96 & 92.48 \\
            MSDFA (k=3,5,7, i=1,3,5) & \textbf{96.01} & \textbf{92.56} \\
        \bottomrule
        \end{tabular}
    }
    \label{table_ablation_msdfa}
\end{table}

\vspace{2mm}\noindent\textbf{Multi-Scale Decoder Fusion with Attention:}
For each ensemble branch in our architecture, we propose using one size larger kernel sizes in MSDFA for both channel and spatial attention. This allows the model to capture more details in the fused feature maps. We analyze this effect in Table~\ref{table_ablation_msdfa}, where we kept the same kernel sizes for all branches in the ensemble compared to a different kernel size for each branch. We found that employing a larger kernel $k=3,5,7$ size than the receptive field size of the ensemble branch $i=1,3,5$ performs better with 96.01\% mF1 and 92.56\% mIoU. The kernel size of 5 for all the ensemble branches achieves 95.99\% mF1 and 92.52\% mIoU, the second-best performance among other combinations.    

\begin{table}[t]
    \centering
    \caption{Ablation study on using different masking probability values on the Levir-CD dataset, results are in \%, and $\delta$ is the masking ratio.}
    \resizebox{0.85\linewidth}{!}{%
        \begin{tabular}{r|cccccc}
        \toprule
            Metric & $\delta=0.1$ & $\delta=0.2$ & $\delta=0.3$ & $\delta=0.4$ & $\delta=0.5$ & $\delta=0.6$ \\
        \midrule
            $mF_{1}$ & 95.94 & 95.97 & \textbf{96.01} & 95.97 & 95.87 & 95.85 \\
            $mIoU$ & 92.43 & 92.47 & \textbf{92.56} & 92.49 & 92.32 & 92.27 \\
        \bottomrule
        \end{tabular}
    }
    \label{table_masking}
\end{table}

\vspace{2mm}\noindent\textbf{Cross-Entropy Masking Ratio:}
Cross-entropy masking balances the impact of a highly imbalanced class ratio in change detection datasets. However, finding a reasonable masking ratio is important for optimal network performance. Higher values of masking can lead to removing too much information for the network to learn, while low values can lead to cross-entropy masking performing equivalent to cross-entropy. We show an analysis of different masking ratios in Table~\ref{table_masking} and found that masking probability $\delta$ equal to 0.3 is optimal. 

\begin{table}[t]
    \centering
    \caption{Ablation study using different losses for class imbalance on the Levir-CD dataset, results are in \%.}
    \resizebox{0.7\linewidth}{!}{%
        \begin{tabular}{r|cccc}
        \toprule
            Method & $mF_{1}$ & $mIoU$ \\
        \midrule
            BCE ($=0.7$) + Dice ($=0.3$) & 95.81 & 92.21 \\
            Focal ($\alpha=0.5$, $\gamma=2.0$) & 95.46 & 91.61 \\
            Weighted-BCE ($w_0=1.0$, $w_1=20.0$) & 92.64 & 86.97 \\
            Weighted-BCE ($w_0=0.3$, $w_1=0.7$) & 95.75 & 92.10 \\
            CEM ($\delta=0.3$) & \textbf{96.01} & \textbf{92.56} \\
            
        \bottomrule
        \end{tabular}
    }
    \label{table_losses}
\end{table}
\vspace{2mm}\noindent\textbf{Losses for Class Imbalance:}
We compared cross-entropy masking (CEM) with other losses widely used for class imbalance, including focal loss~\cite{focalloss}, a combination of binary cross-entropy (BCE) and dice loss~\cite{losses}, and weighted binary cross-entropy~\cite{losses} in Table~\ref{table_losses}. These loss functions adjust the weights of the minority class to penalize it more for false detections. This results in overfitting and even worse performance, as shown in weighted-BCE with equal to 1.0 and $w_1$ equal to 20.0, accounting for the exact class-imbalance in Levir-CD dataset. In weighted-BCE, $w_0$ is the unchanged class and $w_1$ is the changed class. Compared to these, CEM outperforms BCE and other loss functions for class-imbalance. A combination of BCE and dice loss is the second best with 95.81\% mF1 and 92.21\% mIoU.

%Weighted-BCE ($w_0=0.7$, $w_1=1.0$) & 95.96 & 92.46 \\
            
\section{Conclusion}
This paper presents a new architecture to learn change patterns at different scales by using an ensemble of multi-receptive fields. We propose an ensemble of spatial-temporal feature enhancement (STFE), decoder, and multi-scale decoder fusion with attention (MSDFA) to effectively capture local and global contextual change patterns in bi-temporal images. The proposed novel cross-entropy masking (CEM) loss achieved large gains over other losses suggested for class imbalance. Our method demonstrated significant performance gains that outperformed recent remote sensing change detection (RSCD) state-of-the-art (SOTA) methods. 
\label{conclusion}

\section*{Acknowledgment}
This work was supported by the ARC Research Hub in Intelligent Robotic Systems for Real-Time Asset Management (IH210100030).
{
    \small
    \bibliographystyle{ieeenat_fullname}
    \bibliography{main}
}

\end{document}